# Closed-Loop Policies for Operational Tests of Safety-Critical Systems

Jeremy Morton*, Tim A. Wheeler*, and Mykel J. Kochenderfer

*Abstract*—Manufacturers of safety-critical systems must make the case that their product is sufficiently safe for public deployment. Much of this case often relies upon critical event outcomes from real-world testing, requiring manufacturers to be strategic about how they allocate testing resources in order to maximize their chances of demonstrating system safety. This work frames the partially observable and belief-dependent problem of test scheduling as a Markov decision process, which can be solved efficiently to yield closed-loop manufacturer testing policies. By solving for policies over a wide range of problem formulations, we are able to provide high-level guidance for manufacturers and regulators on issues relating to the testing of safety-critical systems. This guidance spans an array of topics, including circumstances under which manufacturers should continue testing despite observed incidents, when manufacturers should test aggressively, and when regulators should increase or reduce the real-world testing requirements for safety-critical systems.

## I. INTRODUCTION

Trust must be established in safety-critical systems prior to their widespread release. Establishing confidence is often difficult due to vast numbers of operational states and infrequent failures. For example, showing that autonomous vehicles are as safe as humans requires performing better than the estimated 530 thousand miles driven between police-reported crashes in the United States, and an estimated 100 million miles driven between fatal crashes [1]. Real-world testing that is too aggressive may yield hazardous events that diminish confidence in system safety. Each manufacturer must devise a testing strategy capable of providing sufficient evidence that their system is safe enough for widespread adoption. Automotive manufacturers can build confidence by conducting test drives on public roadways and make the safety case based on the frequency of observed hazardous events like disengagements and traffic accidents. A manufacturer that is reluctant to test their product may forfeit opportunities to identify and address shortcomings, and may ultimately not be able to compete in the market.

The fundamental tension between the desire to thoroughly test a product and the urgency to forgo further testing in favor of bringing the product to market is not unique to safety-critical systems. Prior work has studied this dilemma in the context of software engineering, where developers must determine whether to delay commercial release in order to devote more time to fixing bugs in their product [2]–[5]. For example, Chávez develops a decision-analytic rule for when to halt software testing. The rule relies on a Bayesian estimate of the bug discovery rate, and weighs costs associated with immediate or delayed bug fixing, loss of market position, and customer dissatisfaction [2]. Littlewood and Wright derive two Bayesian stopping rules for testing safety-critical software, showing the advantages of a conservative reliability-based stopping rule [6].

While relevant, these analyses related to software engineering do not directly translate to all safety-critical systems. In software engineering, a developer is often free to release a product whenever they like—any remaining bugs may cause consumer dissatisfaction, but will not be life-threatening. In contrast, a safety-critical physical system risks inflicting serious injury, death, damage to property, or environmental harm. Thus, the decision of when a safety-critical system is safe enough for release will likely be made by policymakers *not* manufacturers, and manufacturers will likely decide how they should test their product in order to maximize their odds of demonstrating its safety.

This work investigates a test scheduling problem for safety-critical systems, which seeks the best testing schedule to maximize the likelihood of reaching confidence in a candidate system's safety. We present a novel method for deriving closed-loop testing policies, defining the problem as a belief-state Markov decision process (MDP), where high reward is obtained when one becomes certain that the system is safe enough for release. The finite-horizon belief-MDP can be solved exactly using dynamic programming. Furthermore, this approach relies on Bayesian beliefs, which allows prior information to shape manufacturers' decisions and alter their real-world testing burden. Several parametric studies shed insight into test scheduling strategies for real-world automotive manufacturers and regulatory bodies. Though inspired by the challenges of testing autonomous vehicles, the model and its implications are more generally applicable to any safety-critical system that generates inexhaustible and measurable adverse events over a time period or test distance. A section on applying the model to autonomous vehicles is included prior to the conclusions.

## II. A BAYESIAN TESTING FRAMEWORK

Modern testing theory for safety-critical systems typically follows a frequentist approach [7]–[9]. This work deviates from much of the previous work by adopting a Bayesian framework, incorporating informative priors, and deriving offline closed-loop policies. Under a Bayesian framework, data is observed and used to incrementally update a belief. The belief over the unknown hazardous event rate is described probabilistically and changes as information is gathered. Under a frequentist framework, data is assumed to come from a repeatable random sample. The frequentist treats the studies as repeatable, whereas the Bayesian treats the studies as fixed.

Bayesian priors may not always provide coverage—that is, an inferred 95 % credible interval is not guaranteed to

The Department of Aeronautics and Astronautics at Stanford University, Stanford, CA 94305, USA. {jmorton2, wheelert, mykel}@stanford.edu. Asterisk denotes equal contribution by the authors.

contain the true value 95 % of the time over infinite repeated experiments [10]. However, the power associated with the inclusion of Bayesian priors lies in its ability to incorporate informative prior information [11].

The Bayesian framework is well suited to clearly formulating initial belief, calculating the updated belief as information is gathered, and thus for deriving closed-loop policies that incorporate the updated belief to inform future testing. The belief and the derived optimal policy are therefore influenced by the assumed form of the prior belief. A frequentist also makes prior assumptions, but these assumptions are not made explicit in the same way they are for the Bayesian approach [12].

## III. HAZARDOUS EVENT RATES

Suppose that the safety of a safety-critical system is judged based on an expected frequency of hazardous events. Hazardous events such as injuries, fatalities, and disengagements are measurable and their frequency of occurrence should be minimized [13]. A hazardous event frequency is the expected number of hazardous events observed when operating over a given distance or duration of time. The ISO 26262 standard, for example, defines automotive safety integrity levels according to expected hazardous events per unit time [14]. This work measures hazardous events per unit distance due to the less ambiguous nature of having to test over a fixed distance. Knowing the average system speed allows one to convert between the two formulations.

The event frequency of a particular hazardous event for a system under test, $\lambda_{\text{sys}}$, is the expected number of hazardous events encountered when the system is operated over a reference test distance $d_{\text{test}}$. The system is considered safe with respect to the hazardous event when there is credible belief that the event frequency is lower than a reference threshold frequency, $\lambda_{\text{ref}}$. Such a reference frequency may be specified by regulators.

Our model assumes that testing is conducted uniformly across the operational design domain. In reality, the number of encountered hazardous events may vary according to the specific portion of the design domain covered during the test, and not all portions of the design domain are as easily tested. Testing that favors certain scenarios over others may draw incorrect conclusions.

Many statistical methods have been proposed for modeling the rate of hazardous events in roadway driving [15]–[18]. Among the most common are Poisson and Poisson-Gamma (negative binomial) models [19]. Wachenfeld used a Poisson distribution to model the probability of $k$ hazardous events when driving $d_{\text{test}}$ kilometers with a vehicle whose event frequency is $\lambda_{\text{sys}}$ as a Poisson distribution [13]:

$$P(k; \lambda_{\text{sys}}) = \frac{1}{k!}\lambda_{\text{sys}}^k e^{-\lambda_{\text{sys}}} \qquad k = 0, 1, 2, \ldots \quad (1)$$

Poisson distributions capture the probability of a given number of independent, non-exhaustive events occurring in a fixed interval of time or distance given only an average rate of occurrence. The hazardous events under question tend to be extremely rare and arise from the stochastic nature of the driving environment in a manner that they can be assumed to be independent. The rarity of hazardous events also justifies the assumption that the hazardous events are non-exhaustive. Vehicles can be quickly repaired or replaced by the manufacturer in the event of a hazardous event. If many hazardous events occur, it is unlikely that the system will ever establish a high level of safety.

Suppose the core objective of validation is to establish a minimum credibility level $C_{\text{ref}}$ that safety has been achieved:

$$P(\lambda_{\text{sys}} \leq \lambda_{\text{ref}}) \geq C_{\text{ref}}. \quad (2)$$

Thus, a belief over the system's event frequency must be established in order to quantify this release criterion.[1]

## IV. DECISION-THEORETIC FORMULATIONS

This section formulates the safety-critical system test scheduling problem using three successively simpler decision making frameworks. The first, the $\rho$POMDP, is perhaps the most naturally applicable framework. A key insight in this work is the reformulation of the test scheduling problem as a belief MDP, thereby allowing for efficient calculation of policies. The third formulation is a lossless transformation of the belief MDP into an MDP, which aids in interpreting the resulting policies.

### A. $\rho$POMDP

A Markov decision process (MDP) models a sequential decision making problem in a stochastic environment. It is defined by the tuple $\langle \mathcal{S}, \mathcal{A}, T, R, H \rangle$. At any time step $t \in [1, H]$, the system is in a state $s$ contained in the state space $\mathcal{S}$, and the agent must select an action $a$ from the action space $\mathcal{A}$. The transition function $T(s' \mid s, a)$ provides the probability of transitioning from state $s$ to state $s'$ when taking the action $a$ and the reward function $R(s, a, s')$ gives the immediate reward for the transition [20]. While originally developed in the context of operations research, MDPs have since been applied to problems in a broad range of fields, including economics and optimal control [21], [22].

A policy $\pi$ for an MDP is a mapping from states to actions, where $\pi(s)$ denotes the action taken at state $s$. Decision makers seek the optimal policy that maximizes the expected accumulated reward $\sum_{t=1}^{H} r_t$, where $r_t$ is the reward for transition $t$. Optimal policies for finite-horizon MDPs can be computed using dynamic programming [23].

Many problems include uncertainty in the current state and must make inferences about the underlying state based on noisy or partial observations. Uncertainty of this form can be handled by partially observable MDPs, or POMDPs. In lieu of a state, POMDP policies operate on probability distributions over states, inferred from the past sequence of actions and observations. These beliefs, denoted $b(s)$, give the probability of being in state $s$. The space of beliefs is often continuous,

---

[1]An alternative to the probability-density based objective, the reliability prediction-based stopping rule $P(\text{no failures in the next } n_0 \text{tests}) \geq C_{\text{ref}}$ has been shown to produce more conservative policies that more adequately draw conclusions about system reliability [6]. This stopping rule can be directly used in place of Eq. (2).

Table I: Problem formulations

| | | |
|---|---|---|
| $\rho$POMDP | $\mathcal{S}$ | hazardous event rate $\lambda_{\text{sys}}$ |
| | $\mathcal{A}$ | number of tests $n$ |
| | $\mathcal{O}$ | number of hazardous events $k$ |
| Belief MDP | $\mathcal{S}$ | Gamma belief over $\lambda_{\text{sys}}$, $\langle \alpha, \beta \rangle$ |
| | $\mathcal{A}$ | number of tests $n$ |
| MDP | $\mathcal{S}$ | cumulative hazardous events and tests, $\langle K, N \rangle$ |
| | $\mathcal{A}$ | number of tests $n$ |

which can make finding an optimal policy for a POMDP intractable [24], [25].

The reward function for a POMDP has the same form as that of an MDP, $R(s, a, s')$. In the test scheduling problem, however, we do not know a system's true safety level. The reward function in such a problem must instead be dependent on the belief, $R(b, a, b')$. The term $\rho$POMDP refers to POMDPs with with belief-dependent rewards that are piecewise linear and convex [26]. Solving $\rho$POMDPs is an active area of research [27].

The test scheduling problem can be framed as a $\rho$POMDP. The state is the system's hazardous event rate, $\lambda_{\text{sys}}$, and the action is the number of tests $n$ of duration $d_{\text{test}}$ the manufacturer conducts. The state is unobserved, and instead must be inferred from the number of hazardous events, $k$, encountered during testing. The transition function for the $\rho$POMDP in the baseline formulation does nothing; the true hazardous event rate $\lambda_{\text{sys}}$ is assumed constant. The state, action, and observation space are listed in Table I.

The reward function contains two objectives: a terminal reward for achieving sufficient credibility in the safety of the system and a penalty for every hazardous event. We use a scalar parameter $\eta \in [0, 1]$ to trade off between the penalty and reward. The belief and observation-dependent reward function is written:

$$R(b, n, k, b') = \eta \, \text{isterminal}(b') - (1 - \eta)k, \quad (3)$$

where the method isterminal is an indicator function that returns one if the successor belief provides sufficient credibility the system is safe and zero otherwise.

### B. Belief-MDP

The $\rho$POMDP formulation is computationally difficult to solve directly. A key insight into simplifying the problem is that the belief can be represented as a Gamma distribution, providing exact and efficient belief updates. Furthermore, the reachable beliefs are all discrete offsets of the parameters of the initial Gamma distribution, thereby resulting in a discrete belief space that can be iterated over using dynamic programming. We show that while the reachable set of beliefs is infinite, optimal policies can be computed by only considering a finite number of reachable beliefs.

*1) Conjugate Prior:* We extend Wachenfeld's work by representing the belief over the expected hazardous event rate $\lambda$ using the Gamma distribution with density:

$$p(\lambda_{\text{sys}}; \alpha, \beta) = \frac{\beta^\alpha}{\Gamma(\alpha)} \lambda_{\text{sys}}^{\alpha-1} e^{-\beta \lambda_{\text{sys}}}, \quad (4)$$

where $\lambda_{\text{sys}} \in (0, \infty)$ and the belief is parameterized by the shape $\alpha > 0$ and the rate $\beta > 0$. The mean and variance under a Gamma distribution are $\alpha/\beta$ and $\alpha/\beta^2$, respectively.

We write a belief over the hazardous event rate as

$$\lambda_{\text{sys}} \sim \text{Gamma}(\alpha, \beta). \quad (5)$$

The Gamma distribution was chosen because it is a conjugate prior to the Poisson distribution. That is, the posterior belief formed by updating a Gamma distribution with an observation produces a new belief that is also a Gamma distribution. If $k$ hazardous events are observed in the course of performing $n$ tests, each with reference test distance $d_{\text{test}}$, the Bayesian posterior belief is updated according to:

$$\lambda_{\text{sys}} \sim \text{Gamma}(\alpha + k, \beta + n). \quad (6)$$

Such a Gamma belief over the mean of a Poisson-distributed random variable is known as a Poisson-Gamma model. These models are generally preferred to Poisson models in cases where data exhibits over-dispersion, as is often the case with automotive accident data [19].

The credibility of the system safety can be quantified according to:

$$P(\lambda_{\text{sys}} \leq \lambda_{\text{ref}}; \alpha, \beta) = \int_0^{\lambda_{\text{ref}}} p(\lambda; \alpha, \beta) \, d\lambda, \quad (7)$$

which can be evaluated with the aid of the incomplete Gamma function. When $\alpha$ and $\beta$ are integer values, the quantity above can be expressed as

$$P(\lambda_{\text{sys}} \leq \lambda_{\text{ref}}; \alpha, \beta) = 1 - e^{-\beta \lambda_{\text{ref}}} \sum_{m=0}^{\alpha-1} \frac{(\beta \lambda_{\text{ref}})^m}{m!}. \quad (8)$$

The mean and variance under a Poisson-Gamma model are $\alpha/\beta$ and $\alpha/\beta^2$, respectively.

It can be shown that if the hazardous event rate is distributed according to a Poisson-Gamma model, then the number of hazardous events encountered during $n$ trials is governed by the negative binomial distribution [28]:

$$P(k; n, \alpha, \beta) =$$
$$\binom{k + n\alpha - 1}{k} \left(\frac{1}{1+\beta}\right)^k \left(\frac{\beta}{1+\beta}\right)^{n\alpha}. \quad (9)$$

For brevity, we denote the distribution above as $\text{NB}(n\alpha, \beta)$. The expected number of hazardous events encountered in $n$ tests is $n\alpha/\beta$.

*2) Formulation:* The test scheduling problem is formulated as a belief MDP. Each state parameterizes a Gamma distribution over the expected hazardous event rate. The state space is thus the Gamma distribution parameters $\langle \alpha, \beta \rangle$, with shape $\alpha > 0$ and rate $\beta > 0$. The actions are still the number of tests to conduct. The transition function is the belief update Eq. (6), and the reward function is:

$$R(\alpha, \beta, n, k) = \eta \, \text{isterminal}(\alpha + k, \beta + n) - (1 - \eta)k, \quad (10)$$

where $\alpha$ and $\beta$ parameterize the initial belief and the method isterminal($\cdot$) is run on the posterior belief. The belief-MDP representation is summarized in Table I. The set of reachable beliefs is countably infinite.

## C. MDP

Intuition is boosted by making a lossless transformation to an MDP whose states consist of the number of tests conducted and the number of hazardous events observed. Optimal policies obtained for the resulting MDP have equivalent counterparts in the belief MDP. Let the state space $\mathcal{S}$ be the set of all tuples $\langle K, N \rangle$, corresponding to the cumulative number of hazardous events $K$ encountered in $N$ total tests. The action space $\mathcal{A}$ is the set of all nonnegative integers corresponding to the number of tests to perform in the next quarter. We first present the baseline MDP formulation, where the MDP state is equivalent to the belief over system safety ($\langle K, N \rangle \Leftrightarrow \langle \alpha, \beta \rangle$). We then explain how this formulation can be extended to account for manufacturer innovation and prior beliefs over system safety.

The reward function for the MDP is still belief-dependent, but is expressed in terms of the current state, number of tests, and number of observed adverse events. The reward function for the baseline problem formulation is given by:

$$R(K, N, n, k) = \eta \, \mathrm{isterminal}(K+k, N+n) - (1-\eta)k. \quad (11)$$

In the test scheduling problem, we seek a policy $\pi^* : \mathcal{S} \to \mathcal{A}$ that prescribes the optimal number of tests to perform in each state.

*1) Maximizing Immediate Reward:* We first demonstrate how to find a policy that maximizes the immediate reward obtained by a manufacturer. We know the optimal test scheduling policy will maximize the expected reward:

$$\pi^*(K, N) = \arg\max_n \mathbb{E}_k \left[ R(K, N, n, k) \right], \quad (12)$$

where $k \sim \mathrm{NB}(nK, N)$.

We know that

$$\mathbb{E}_k[(1-\eta)k] = (1-\eta)n\frac{K}{N}, \quad (13)$$

but computing $\mathbb{E}_k[\eta \, \mathrm{isterminal}(K+k, N+n)]$ requires taking an expectation over a countably infinite set of values for $k$. However, close inspection of Eq. (8) reveals that if $P(\lambda_{\mathrm{sys}} \leq \lambda_{\mathrm{ref}}; K + \tilde{k}, N + n) \leq C_{\mathrm{ref}}$ for some $\tilde{k}$, then $P(\lambda_{\mathrm{sys}} \leq \lambda_{\mathrm{ref}}; K + k, N + n) \leq C_{\mathrm{ref}}$ for all $k \geq \tilde{k}$. Thus, it is possible to calculate $\mathbb{E}_k[\eta \, \mathrm{isterminal}(K+k, N+n)]$ efficiently by iterating until a value of $k$ is encountered such that it is not possible to reach a terminal state.

Finding an optimal policy for each state requires finding a value of $n$ that maximizes the expected reward. However, we cannot just loop over all values of $n$ to find the optimal value, as the number of possibilities are countably infinite. Hence, we desire an upper bound on $n^*$, the optimal amount of tests to perform. Note that the expected reward has a lower bound of zero, as it is always an option to not perform any tests and receive no reward. Furthermore, it is clear that $\mathrm{isterminal}(\cdot)$ has an upper bound of one. Using these properties, we find an upper bound on $n^*$:

$$0 \leq \mathbb{E}_k \left[ \eta \, \mathrm{isterminal}(K + k, N + n^*) - (1-\eta)k \right] \quad (14)$$

$$\leq \eta - (1-\eta)n^* \frac{K}{N} \quad (15)$$

$$\implies n^* \leq \frac{\eta}{1-\eta} \frac{N}{K}. \quad (16)$$

*2) Baseline Test-Scheduling Problem:* We now formulate a decision problem under which testing decisions can be made over the course of multiple quarters. In this *sequential* decision problem, we must consider the possible rewards that can be received far into the future, not just immediate rewards. If we wish to derive a testing schedule over the course of $T$ quarters, then we must solve for policies that are functions of both the state $\langle K, N \rangle \in \mathcal{S}$, as well as the current time step $t \in \{1, 2, \ldots, T\}$.

For each state and time step, the aim is to determine the optimal utility, which captures the expected reward that will be received at all future time steps through following the optimal policy. According to the Bellman equation, the optimal utility $U^*(K, N, t)$ will satisfy the recurrence [23]:

$$U^*(K, N, t) = \max_n \mathbb{E}_k [R(K, N, n, k) + \gamma U^*(K + k, N + n, t + 1)], \quad (17)$$

where $\gamma \in [0, 1]$ is the discount factor, a parameter that discounts rewards received in future time steps. We can limit the number of values of $n$ that we must consider by using the upper bound derived in Eq. (14). Furthermore, for any terminal state $\langle K, N \rangle$ it is assumed that $U^*(K, N, t) = 0$ for any time step $t$.

The optimal policy for the terminal time step $T$ will be equivalent to the optimal policy under the immediate reward model, as it is not possible to attain any subsequent rewards. Thus, the optimal policy and associated utility can be found for each state $\langle K, N \rangle \in \mathcal{S}$ at the terminal time step, and then a single backward sweep through time can be performed using Eq. (17) to calculate optimal utilities and policies at all preceding time steps.

In order to calculate the expected utility in future time steps $\mathbb{E}_k[U^*(K + k, N + n, t + 1)]$, we must again take an expectation over a countably infinite number of hazardous events. However, it can be shown that if $U^*(K + \tilde{k}, N + n, t + 1) = 0$ for some non-terminal state $\langle K + \tilde{k}, N + n \rangle$, then $U^*(K + k, N + n, t + 1) = 0$ for all $k \geq \tilde{k}$ (see appendix). Thus, we only consider the utility of the future states up to the point that a state is encountered that is both non-terminal and has zero utility.

*3) Innovation Bonus:* The baseline test scheduling problem described above does not allow for the safety of a system to improve over time through research and development. Hence, we propose a modification to the baseline problem formulation that allows the safety of a system to evolve stochastically over time through an innovation bonus. To accommodate this change, the state space is augmented with an additional integer-valued innovation term $\beta_I$ that alters the expected number of hazardous events that will be encountered in future testing.

Between subsequent quarters, the change in innovation level $\Delta \beta_I$ is drawn according to a categorical distribution over integer values and scaled linearly according to the number of tests performed. The scaling of innovation with tests models the ability of manufacturers to identify and remedy problems with their system in the course of real-world testing. For a given state $\langle K, N, \beta_I \rangle \in \mathcal{S}$, the number of hazardous events encountered in $n$ tests is distributed according to a negative binomial distribution that accounts for the current innovation level:

$$k \sim \text{NB}(nK, N + \beta_I). \tag{18}$$

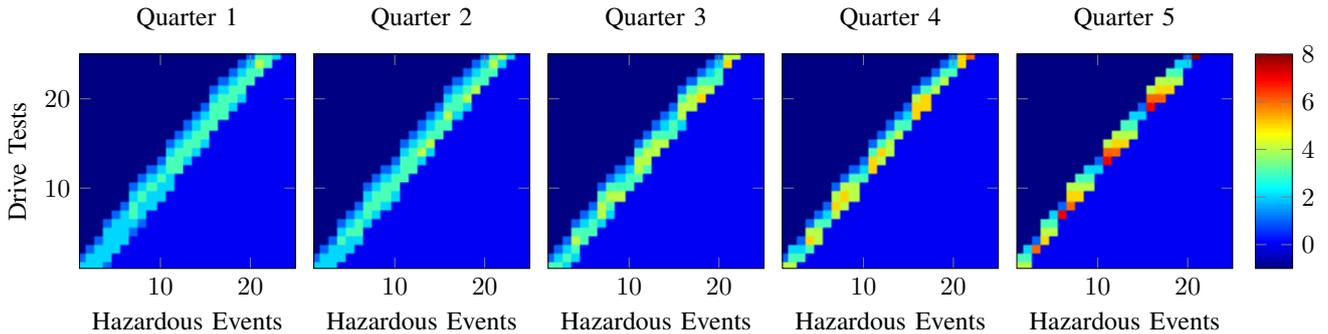

Figure 1: Example five-quarter testing policy with parameters shown in Table II. The innovation distribution contains probabilities of $P(\Delta\beta_I = 0) = 1/2$, $P(\Delta\beta_I = 1) = 1/4$, and $P(\Delta\beta_I = 2) = 1/4$. Each square in a given plot corresponds to a state $\langle K, N \rangle$, and the color of the square corresponds to the number of tests that the policy prescribes for that state. States corresponding to lower observed hazardous event rates can be found in the top left of each plot, and states corresponding to higher observed rates can be found in the bottom right. As such, terminal states can be found in the dark blue regions that occupy the top left of each plot.

Thus, while the innovation bonus cannot alter the number of hazardous events that have already occurred, it can make further hazardous events more or less likely depending on the value of $\beta_I$. Positive innovation bonuses decrease the expected number of hazardous events and negative innovation bonuses increase the expected number. In the absence of an innovation bonus, the distribution over future hazardous events is the same as the distribution in the baseline test scheduling problem. Hence, the exact form of the innovation distribution will guide manufacturer decisions about when they should perform tests, as it can influence their beliefs about whether the safety of the system will change relative to its observed performance.

*4) Prior Belief:* The test scheduling formulations presented above have assumed the absence of any relevant information that might influence a manufacturer's belief about the safety of the system under test. However, in testing a real-world system, there may exist several relevant sources of information that would guide expectations about how that system will perform in future tests. For example, a manufacturer may wish to incorporate confidence based on simulation-based testing, or confidence due to the system under test being a variant of a previously certified system.

Suppose we have relevant experience that shows a hazardous event frequency $\mu$, and we would like to use it to construct a prior for our belief over $\lambda_{\text{sys}}$, the hazardous event frequency of interest. As a prior belief over $\lambda_{\text{sys}}$, we can assume it is drawn from a Gamma distribution with mean $\mu$ and user-specified variance $\sigma^2$. A small value of $\sigma^2$ would indicate a strong belief that the true value of $\lambda_{\text{sys}}$ is close to $\mu$, while a large value of $\sigma^2$ would indicate that $\mu$ may not be very informative about the true value of $\lambda_{\text{sys}}$.

We introduce the values $\alpha_0$ and $\beta_0$, which will parameterize the prior distribution over $\lambda_{\text{sys}}$:

$$\lambda_{\text{sys}} \sim \text{Gamma}(\alpha_0, \beta_0). \tag{19}$$

Using the mean and variance of the Poisson-Gamma model, we find the following relations [29]:

$$\mu = \frac{\alpha_0}{\beta_0}, \quad \sigma^2 = \frac{\alpha_0}{\beta_0^2}. \tag{20}$$

Rearranging and solving, we find that $\beta_0 = \mu/\sigma^2$ and $\alpha_0 = \mu\beta_0$. By performing tests, we can incrementally update our belief over the hazardous event frequency of interest through the posterior distribution:

$$\lambda_{\text{sys}} \sim \text{Gamma}(\alpha_0 + K, \beta_0 + N). \tag{21}$$

By incorporating this prior belief, the baseline test scheduling problem is modified in two ways. First, the reward function changes, as the inclusion of prior knowledge alters the amount of experience required to reach a terminal state:

$$\text{isterminal}(K, N) = \\ \mathbf{1}\left\{P(\lambda_{\text{sys}} \leq \lambda_{\text{ref}}; K + \alpha_0, N + \beta_0) \geq C_{\text{ref}}\right\}. \tag{22}$$

Furthermore, prior belief can alter the distribution over the number of hazardous events that may be encountered during testing. The distribution over hazardous events is given by:

$$k \sim \text{NB}\left(n(K + \alpha_0), N + \beta_0 + \beta_I\right), \tag{23}$$

where $\beta_I$ is the innovation bonus described previously.

An example five-quarter sequential policy is given in Fig. 1. The innovation distribution is described in the figure caption, and the remaining problem parameters can be found in Table II. In each policy plot, the terminal states can be found in the dark blue region of the upper left corner and the states where no testing occurs can be found in the lighter blue region in the bottom right corner. Note that testing tends to occur in states adjacent to the terminal states. Furthermore, fewer tests tend to occur in each state at early time steps, which can be attributed to the influence of the discount factor.

## V. PARAMETRIC STUDIES

In the following sections, we present a series of takeaways that result from studying policies obtained through the procedures outlined above. These takeaways are meant to provide insight to manufacturers as they grapple with high-level decisions about system testing. We hope the following analysis can provide guidance to both manufacturers and regulators about how the testing and certification of safety-critical systems can be accelerated in a safe manner.

Table II: Problem parameters for policy in Fig. 1

| $C_{\text{ref}}$ | $\lambda_{\text{ref}}$ | $\eta$ | $\gamma$ | $\mu$ | $\sigma^2$ |
|---|---|---|---|---|---|
| 0.95 | 1 | 0.95 | 1 | 0.5 | 0.1 |

Table III: Minimum reward ratio required for any testing when $\frac{K}{N} > \lambda_{\text{ref}}$.

| $C_{\text{ref}}$ | 0.90 | 0.95 | 0.99 |
|---|---|---|---|
| $\left(\frac{\eta}{1-\eta}\right)_{\min}$ | $3.50\times 10^2$ | $1.80\times 10^3$ | $2.52\times 10^4$ |

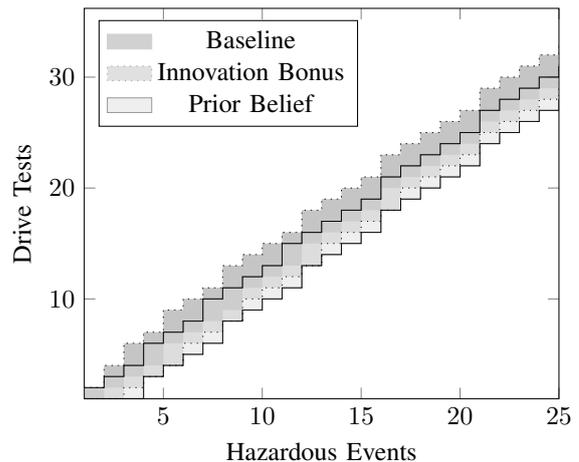

Figure 2: Regions of the state space where testing occurs for baseline problem, baseline with innovation bonus, and baseline with prior belief.

*A. If only concerned about immediate reward, a manufacturer should avoid tests when the observed hazardous event rate exceeds $\lambda_{ref}$.*

Of particular interest in the test scheduling problem are scenarios where a safety-critical system has *not* demonstrated the requisite safety level in prior testing. For example, consider a system that has undergone one test and experienced two hazardous events. If we assume that $\lambda_{\text{ref}} = 1$, then the observed hazardous event frequency for this system exceeds the reference value. Such an outcome is troubling; while outside factors may have contributed to the observed hazardous events, a hypothesis that the system is sufficiently safe now requires heavy skepticism. Hence, it is worth pondering the circumstances under which it may be optimal to continue testing a system with a subpar safety record.

If a manufacturer is only concerned about immediate reward then they completely discount reward that can be received in subsequent quarters. Thus, they must weigh the reward for achieving strong belief that safety requirements are met against the penalty associated with hazardous events that occur during testing. By modifying the structure of the reward function, we can incentivize or discourage testing in different regions of the state space. In Table III, we present the minimum reward ratio ($\frac{\eta}{1-\eta}$) required in order for any tests to be administered when the observed hazardous event rate exceeds the reference value $\lambda_{\text{ref}} = 1$ (i.e. when $K/N > 1$). Results are presented for credibility levels of 90 %, 95 %, and 99 %.

From the table, we see that the reward associated with reaching a terminal state must be at least hundreds to thousands of times larger than the cost associated with a hazardous event in order for any testing to be performed in cases where a safety-critical system has not met the reference safety standard in previous testing. It is unrealistic to assume that any manufacturer would care so little about hazardous events, in particular fatalities, that they would have such a lopsided reward function. Therefore, we conclude that manufacturers should not make an all-at-once attempt to prove the safety of a system if previous tests have shown the system to have an inadequate safety record.

*B. Expected innovation and prior beliefs can justify more testing.*

In contrast with manufacturers who are myopic in pursuing immediate rewards, the sequential test scheduling problem allows a manufacturer to reason about how their system's safety record may improve in future testing. The two primary mechanisms that can justify such optimism are (1) expectations of improved system performance through innovation and (2) prior beliefs that suggest the system is likely to operate safely in the future. In practice, innovation or prior beliefs may make it optimal to perform tests in certain states $\langle K, N \rangle \in \mathcal{S}$ where testing would not occur in the baseline sequential test scheduling problem.

This more aggressive testing strategy is shown in Fig. 2. The *baseline* region of the plot illustrates the states where it would be optimal to perform any tests during the first of five quarters in the baseline problem associated with Table II. Furthermore, the plot shows how the testing region varies with either the inclusion of an innovation bonus described in the caption of Fig. 1 or the incorporation of a prior belief with $\mu = 0.5$ and $\sigma^2 = 0.1$. Both sources of optimism cause the region of the state space where testing occurs to extend downward, which signifies that testing is occurring in states associated with higher observed hazardous event rates. An additional effect of incorporating prior experience can be noticed as well, as the upper bound of the testing region is decreased. In this case, the prior experience inspires a stronger belief that the system is sufficiently safe, which reduces the testing burden and increases the number of terminal states.

Figure 2 provides illustrative examples of scenarios where manufacturers should be willing to tolerate higher hazardous event rates and continue testing. With the innovation bonus, manufacturers expect their system to become safer over time, and hence assume that the hazardous event rates encountered in future testing will be lower. In contrast, relevant experience can inspire optimism that the system is in fact already safer than what has been observed so far, and that higher hazardous

Table IV: Testing strategies for various discount factors

| $\gamma$ | 0.5 | 0.75 | 1.0 |
|---|---|---|---|
| Average number of tests per state | 3.20 | 2.44 | 2.16 |
| Fraction of states with testing | 0.09 | 0.11 | 0.13 |

event rates may be anomalous and likely to decrease in future tests.

*C. A patient manufacturer tests incrementally, but may tolerate higher hazardous event rates.*

Manufacturers are motivated to be the first to market. They must balance the risk of aggressive testing against the relative safety of letting the research and development team improve system safety over a longer time horizon. In the sequential test scheduling problem, the balance between this desire for immediate over long-term rewards is controlled by a discount factor $\gamma$. In the limit as $\gamma \to 0$, all future rewards are neglected, and the policy in each quarter should be identical to the policy for a manufacturer that only cares about immediate rewards. However, as the value of the discount factor approaches one, future rewards are valued as much as immediate rewards, and the optimal testing strategy for a manufacturer can change considerably.

The change in testing strategy as $\gamma \to 1$ manifests itself in two ways, both of which are most apparent in early time steps. First, in states where testing does occur, the number of tests tend to decrease. Second, there is an increase in the number of states where testing occurs. Table IV shows the average number of tests performed in each state and the fraction of nonterminal states where testing occurs as the discount factor is varied between 0.5 and 1 for the first quarter of a test scheduling problem that otherwise is defined by the parameters given in Table II. While these results are for a single problem instantiation, they are illustrative of a trend that is observed in sequential test scheduling problems with a variety of reward structures, innovation distributions, and forms of relevant experience.

The first change in testing strategy, where fewer tests are performed in each state, comes from tests being delayed until the end of the finite time horizon. Reward is not discounted, so there is no penalty in delaying a large portion of testing until the final quarter.

The second change in testing strategy, that a higher discount factor makes it optimal to perform tests in *more* states, is somewhat counterintuitive. The states where testing becomes optimal as $\gamma \to 1$ are associated with higher hazardous event rates. As such, it is undesirable to perform too many tests in these states, which means it may take several quarters of testing before a manufacturer can achieve a sufficiently strong belief that their system is safe. Because this terminal reward signal can be quite delayed for states associated with a higher hazardous event rate, testing is only worthwhile if a manufacturer is patient and does not heavily discount future rewards relative to immediate rewards.

There is extensive interplay between the effect of manufacturer patience and the optimism engendered by the innovation bonus and prior beliefs. In states associated with higher hazardous event rates, expected innovation or outside information can spur a manufacturer to believe that hazardous event rates could be lower in future tests. However, depending on the strength of this belief, manufacturers may only find it worthwhile to perform a limited number of tests. Hence, such restrained testing is most justified in cases where a delayed terminal reward for making it to market still outweighs the costs of hazardous events that occur during testing.

*D. In the sequential decision problem, it can be optimal to perform tests even though the observed hazardous event rate exceeds $\lambda_{ref}$.*

Manufacturers that only value immediate rewards do not perform incremental testing or consider how the performance of their system may improve over time. Instead, they only use the observed performance of their system in order to predict how it will fare in future testing. In previous sections, we have shown how expected improvement, relevant experience, and the discount factor all interact to guide manufacturer decisions when system testing is performed over the course of several quarters instead of all at once. Furthermore, we have shown how these components of the sequential test scheduling problem can inspire manufacturers to be more aggressive about testing in states associated with higher hazardous event rates.

Under certain circumstances, it is optimal to perform testing even though previous tests have yielded an observed hazardous rate that exceeds $\lambda_{ref}$. The gray curves in Fig. 3 show the maximum observed hazardous event rate, $(K/N)_{max}$, under which it would be optimal to perform any testing for the testing policies shown in Fig. 1. These curves illustrate multiple instances where it is optimal to perform testing even though the hazardous event rate exceeds $\lambda_{ref} = 1$. The values of $(K/N)_{max}$ are plotted against the total number of tests performed, meaning that hazardous event rates associated with a smaller number of tests exhibit a higher level of variance. Hence, it is clear that manufacturers can choose to be more aggressive with testing in cases where there is more uncertainty in a system's estimated safety level.

While we have shown that it *can* be optimal to perform tests in cases where the observed hazardous event rate is higher than the reference standard, it is worth stressing that such circumstances are rare. The black curves in Fig. 3 illustrate optimal release decisions for the baseline problem associated with Table II (i.e. the same problem without the innovation bonus or relevant experience). These curves show that under such circumstances it is never optimal to perform testing when the hazardous event rate exceeds $\lambda_{ref}$. Thus, in many cases it is not optimal to continue testing when previous performance has shown a system to be insufficiently safe. However, the following factors would make it more likely that it is worthwhile to continue testing a safety-critical system that has already experienced a high number of hazardous events:

1) High uncertainty in the hazardous event rate (few tests have been performed).
2) Many more quarters remaining for testing.
3) High reward for reaching a terminal state relative to penalty for hazardous events.

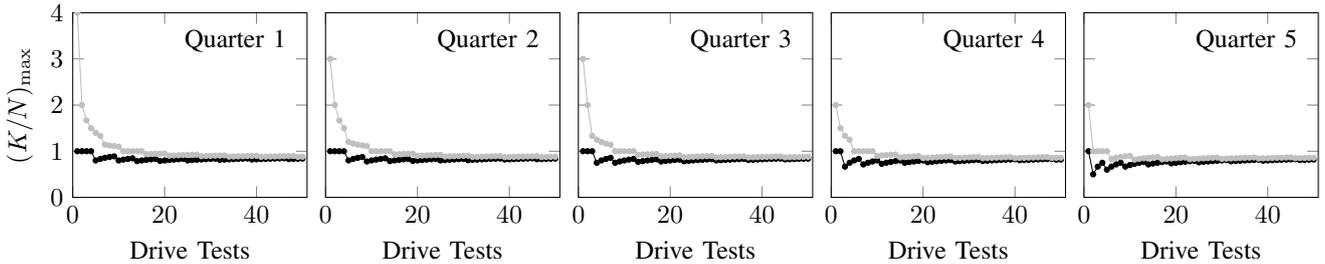

Figure 3: Maximum observed hazardous event rate for which safety-critical systems would be released as a function of total number of tests performed. The gray curves show values for the policies shown in Fig. 1 with innovation probabilities of $P(\Delta\beta_I = 0) = 1/2$, $P(\Delta\beta_I = 1) = 1/4$, and $P(\Delta\beta_I = 2) = 1/4$. The black curves show values for the optimal policies of the associated baseline problem without innovation.

4) High likelihood of improvement through innovation.
5) Prior encodes strong belief that the system is safer than previous tests suggest.
6) High discount factor ($\gamma \approx 1$).

*E. Incorporating prior beliefs can change the amount of required testing.*

Aside from influencing testing decisions for manufacturers, prior beliefs can also provide insights about how regulators may be able to alleviate the immense real-world testing burden necessary to prove the safety of advanced safety-critical systems such as automated vehicles. Rather than making the safety case entirely through tests, it may be possible to augment testing experience with other sources of information, such as performance in simulation.

The stronger the belief that the relevant experience is predictive of the safety level of the system under test, the stronger it will bias the estimate of the hazardous event rate for that system. This trust in the predictive value of the experience is encoded in its estimated variance, $\sigma^2$, and will need to be determined on a case-by-case basis. In Fig. 4, we see the effect that different values of $\mu$ and $\sigma^2$ have on the real-world testing burden required to prove system safety. The top three plots show the change in the number of terminal states in a $50 \times 50$ state space ($K \in \{1, \ldots, 50\}$, $N \in \{1, \ldots, 50\}$) for three values of $\mu$ and a range of variance values when $C_{\text{ref}} = 0.95$ and $\lambda_{\text{ref}} = 1$. Reaching a terminal state means no additional testing is required to prove system safety. Thus, an increase in the number of terminal states means that the testing burden is reduced, while a decrease in the number of terminal states signifies that even more testing should be performed to prove system safety.

Three distinct behaviors can be observed for different values of $\mu$. In the top left plot, when $\mu = 0.5$, we see what is deemed *Type* 1 behavior, where at low variance values there is an increase in the number of terminal states, with the change in terminal states trending toward zero as the variance becomes large and the prior becomes non-informative. In contrast, in the top right plot, when $\mu = 1$ we see *Type* −1 behavior, where the number of terminal states decreases for all variance values and the change tends toward zero for large variances. In the top middle plot, when $\mu = 0.9$, we see *Type 0* behavior, which like *Type* 1 behavior exhibits an increase in the number of terminal states at low variance values and little net effect at high variances. Interestingly, however, at intermediate variance values it shows a decrease in the number of terminal states (and hence an increased testing burden). This behavior arises because $\mu$ is close to $\lambda_{\text{ref}}$, and at moderate variance levels the uncertainty is high enough that we must consider that the hazardous event rate for the system may actually exceed the reference standard.

The bottom three plots of Fig. 4 show that the testing burden can change as $\mu$ and $C_{\text{ref}}$ vary. For $C_{\text{ref}} \geq 0.95$, we see that for $\mu < \lambda_{\text{ref}}$ there is either *Type* 1 or *Type* 0 behavior, while for $\mu > \lambda_{\text{ref}}$ there is always *Type* −1 behavior. This provides two practical takeaways for regulators. First, if a manufacturer can demonstrate through relevant experience that their hazardous event rate may be lower than the reference standard, then it is possible that their testing burden can be reduced, although this depends on the assumed predictive power of such experience. Second, if relevant experience suggests that a safety-critical system may not be sufficiently safe, then there should *always* be an enhanced real-world testing burden on the manufacturer.

*F. When should the most testing be performed?*

We have studied the conditions necessary to make it worthwhile for a manufacturer to perform *any* tests, but we have not yet addressed a related question: under which conditions should a manufacturer perform the most testing? In particular, it is worth considering which observed hazardous event rates should make a manufacturer most motivated to aggressively test their safety-critical system. To provide a clearer answer to this question, Fig. 5 plots the optimal number of tests to perform in a terminal time step against the observed hazardous event rate with $\eta = 0.99$ and $C_{\text{ref}} = 0.95$. The high terminal reward and lack of opportunity for future tests in this problem provides a strong incentive for performing a large amount of immediate testing. With a large amount of testing performed, it is easier to see a trend in the results.

The points in Fig. 5 are colored according to the number of tests already performed, and hence the color can be seen to represent the variance in the hazardous event rate estimate. The lighter points therefore exhibit highest variance, while the darker points have the lowest variance. This figure shows that the largest number of tests tend to occur in states with low hazardous event rates and high variance. Hence, manufacturers should be most aggressive about testing in circumstances when their observed hazardous event rate is low, but they have not yet performed sufficient testing to credibly demonstrate that their system meets the reference safety standard.

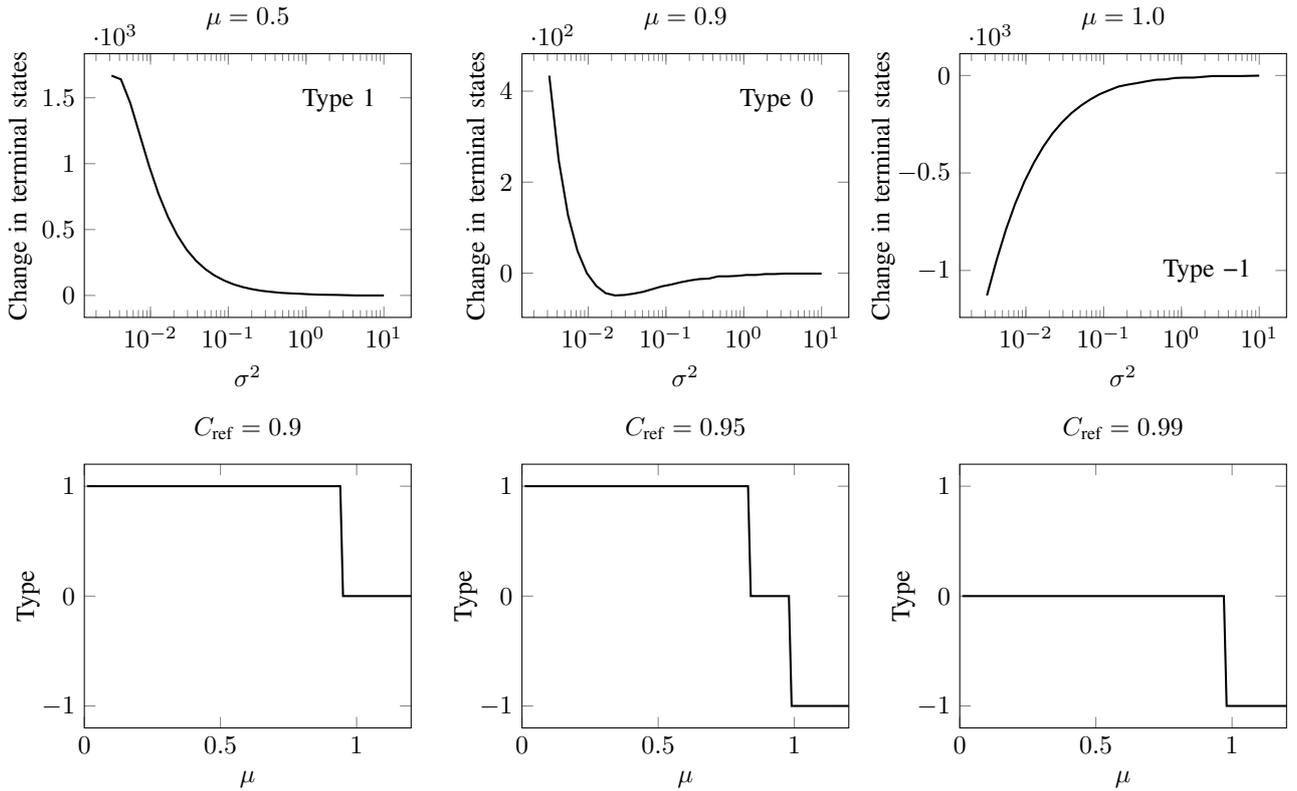

Figure 4: Visualization of the change in number of terminal states due to incorporation of relevant experience. The top three plots show that change in number of terminal states in a $50 \times 50$ state space as a function of mean and variance. In the bottom three plots, we see the changes in testing burden as a function of mean $\mu$ and credibility level $C_{\text{ref}}$.

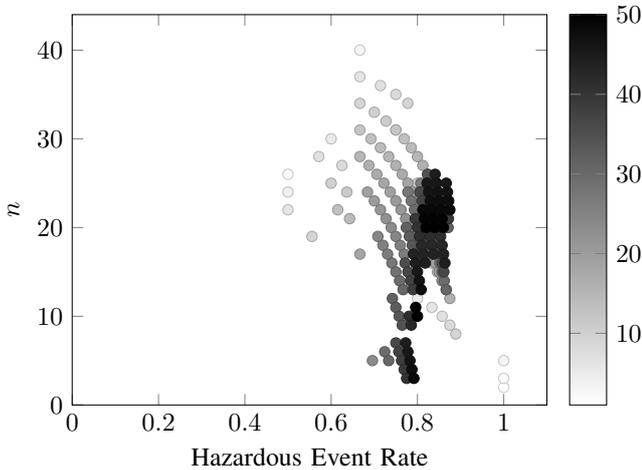

Figure 5: Optimal number of tests to perform as a function of observed hazardous event rate with $\eta = 0.99$ and $C_{\text{ref}} = 0.95$. The color of each point represents the number of tests already performed.

Table V: Hazardous event frequencies [32], [33].

| | |
|---|---|
| Disengagement Rate | 0.121 per 1000 km |
| Rate of Reported Collisions | 12.5 per 100 million km |
| Fatality Rate | 0.70 per 100 million km |

## VI. APPLICATION TO AUTONOMOUS VEHICLES

The development of autonomous vehicles presents unprecedented challenges in safety validation [7], [8], [13], [30]. Koopman and Wagner outline several specific challenges in autonomous vehicle testing and validation, including the infeasibility of complete testing, the inability to rely on humans to intervene, and the difficulty of testing systems that may be stochastic or learning-based [8]. Perhaps principal among these challenges is that the space of driving situations is so large that sufficient testing to ensure ultra-dependable system operation is not possible. In showing that the fatality rate of an active safety system is less than one fatality per $10^9$ hours of operation, one must conduct a multiple of $10^9$ operation hours of testing in order to achieve statistical significance [31].

The frequency of common hazardous events vary greatly in magnitude. As Table V illustrates, the disengagement rate differs from the fatality rate by five orders of magnitude. Waymo has covered the largest test distance to date, with over four million test miles [32], but even such distances are merely sufficient for determining the likelihood of the most common hazardous events. Proper estimates for the fatality rate require driving a multiple of the average 70 million kilometers between collisions for human drivers [33]. Unfortunately, the rarest events are often the most critical in determining whether to release an autonomous vehicle, leading to significant uncertainty in the safety of a system even after millions of miles of testing.

The size and scope of the real-world testing burden that manufacturers will face is currently unknown. Given the rarity of the most extreme hazardous events, such as fatalities, building a case that a system meets safety standards based on observed hazardous event rates would require an infeasible amount of real-world testing [7], [13], [34]. Recent research has proposed methods aimed at reducing the amount of real-world testing required from manufacturers. Such methods include testing in virtual and hardware-in-the-loop simulations, limiting the scope of real-world testing to safety-critical scenarios, and using threat measures from near-collisions to quantify system safety [1], [30], [35]–[37].

Successful application of the proposed model to autonomous driving requires identifying suitable parameters. Although exact values have not been established to date, we can use hazardous event rates under human driving as preliminary values for the reference frequency $\lambda_{\text{ref}}$ [30], [34]. The granularity of the test distance $d_{\text{test}}$ is a free parameter. The minimum credibility $C_{\text{ref}}$ is also a free parameter, but is likely to be at a standard level such as 95 % or 99 %. The reward tradeoff parameter $\eta \in [0,1]$ can also be adjusted by the manufacturer. The value of $\eta$ would vary according to the hazardous event of interest, with values close to one signifying that the reward associated with reaching a terminal state far outweighs the cost of a single hazardous event. Trends in innovation for batteries, sensors, and related technologies, as well as aviation safety records can potentially be used to estimate or bound the innovation bonus.

A more elusive parameter is the prior belief over the system's safety, parameterized by a Gamma distribution via $\alpha$ and $\beta$. The choice of prior can significantly affect the result—as is desired—and any serious use of the model must be able to justify the prior used. The Gamma model does not allow for encoding a uniform prior. Rather, a mean and variance must be selected that adequately capture the manufacturer's prior belief. Possible sources by which such a mean and variance can be justified include previously tested systems similar to the system under test, the system's performance on hazardous events with higher encounter frequencies, and the system's performance in simulation.

## VII. CONCLUSIONS

This work introduced a method for deriving closed-loop testing strategies for safety-critical systems. The problem is formulated as a finite-horizon Markov decision process, where policies map a system's observed safety record to a recommendation for how many tests should be carried out in the next quarter. By performing parametric studies on a set of optimal testing policies, we arrive at a series of insights that can guide manufacturers. We discussed the range of assumptions that manufacturers must make in order to continue testing their product in circumstances where previous testing has shown it to be insufficiently safe. Additionally, we showed how regulators can use information obtained from sources other than operational testing to reduce the testing burden for a safety-critical system. Finally, we demonstrated that manufacturers should perform the most testing in cases where their observed hazardous event rate is low, but there is high uncertainty in the estimate.

The initial goal of this project is to provide manufacturers with high-level takeaways about relevant considerations relating to testing safety-critical systems, not to derive an optimal testing policy for any particular system. As such, future work must be dedicated to modifying the problem formulation in areas where it lacks the appropriate level of fidelity. In particular, further work is required in devising methods for determining quantities such as the innovation bonus, discount factor, and prior beliefs in a principled manner. Additionally, our analysis does not account for the cost of performing tests, but manufacturers may wish to account for such costs in deriving their own testing policies. To facilitate further analysis, the code for this project has been released to the public, and can be found at https://github.com/sisl/av_testing.

## APPENDIX A
## PROOF THAT SUBSEQUENT STATES HAVE ZERO UTILITY

Here, we prove that if $U^*(K,N,t) = 0$ for some state $\langle K,N \rangle$, then $U^*(K+\ell,N,t) = 0$ for all positive integers $\ell$. We limit the proof in this section to $\ell = 1$ and the terminal time step $T$, but it follows inductively that the same analysis holds for any greater $\ell$ and for all time steps $t$. We know that the optimal utility at the terminal time step satisfies:

$$U^*(K,N,T) = \max_n \mathbb{E}_k[\eta \, \text{isterminal}(K+k, N+n) - (1-\eta)k], \quad (24)$$

where $k \sim \text{NB}(nK, N)$. If we encounter a state $\langle K,N \rangle$ where $U^*(K,N,T) = 0$, then we know that

$$\mathbb{E}_k[\eta \, \text{isterminal}(K+k, N+n)] \leq (1-\eta)n\frac{K}{N} \quad (25)$$

for all $n$.

Now consider a state $\langle K+1, N \rangle$. From Eq. (8):

$$\mathbb{E}_{k \sim \text{NB}(n(K+1),N)}[\text{isterminal}(K+k+1, N+n)] \leq \mathbb{E}_{k \sim \text{NB}(n(K+1),N)}[\text{isterminal}(K+k, N+n)]. \quad (26)$$

We would like to show that

$$\mathbb{E}_{k \sim \text{NB}(n(K+1),N)}[\text{isterminal}(K+k, N+n)] \leq \mathbb{E}_{k \sim \text{NB}(nK,N)}[\text{isterminal}(K+k, N+n)]. \quad (27)$$

For each $n$, we can say that

$$\mathbb{E}_{k \sim \text{NB}(nK,N)}[\text{isterminal}(K+k, N+n)] = \sum_{k=0}^{k^{(n)}} P(k; n, K, N), \quad (28)$$

where $k^{(n)}$ is the largest value of $k$ such that $\text{isterminal}(K + k, N + n)$ takes on a value of one. Hence, the above quantity represents the cumulative distribution function over $k$.

For the negative binomial distribution, the cumulative distribution function is given in terms of the regularized beta function $I_p(\cdot)$ [38]. We have that

$$\mathop{\mathbb{E}}_{k \sim \text{NB}(n(K+1), N)}[\text{isterminal}(K + k, N + n)] = I_p(nK + n, k^{(n)} + 1) \tag{29}$$

and

$$\mathop{\mathbb{E}}_{k \sim \text{NB}(nK, N)}[\text{isterminal}(K + k, N + n)] = I_p(nK, k^{(n)} + 1), \tag{30}$$

where $p = N/(1 + N)$. The regularized beta function has the property that $I_p(a + 1, b) \leq I_p(a, b)$. Thus, we find that the inequality in Eq. (27) holds.

We denote the expected utility of performing $n$ tests as $U^{(n)}(\cdot)$. For each $n$, it follows that

$$U^{(n)}(K + 1, N, T) = \mathop{\mathbb{E}}_k[\eta \, \text{isterminal}(K + k + 1, N + n) - (1 - \eta)k], \tag{31}$$

where $k \sim \text{NB}(n(K + 1), N)$. Using the inequalities in Eq. (26) and Eq. (27), we find that

$$U^{(n)}(K + 1, N, T) \leq \mathop{\mathbb{E}}_k[\eta \, \text{isterminal}(K + k, N + n)] - (1 - \eta)n\frac{K + 1}{N} \tag{32}$$

where $k \sim \text{NB}(nK, N)$. Finally, we use the result from Eq. (25) to find that:

$$U^{(n)}(K + 1, N, T) \leq (1 - \eta)n\frac{K}{N} - (1 - \eta)n\frac{K + 1}{N} \tag{33}$$
$$= -(1 - \eta)\frac{n}{N} \leq 0. \tag{34}$$

Thus, we see that performing any positive number of tests has a negative expected utility, and hence $U^*(K + 1, N, T) = 0$.

## ACKNOWLEDGMENT

This material is based upon work supported by the National Science Foundation Graduate Research Fellowship Program under Grant No. DGE-114747 and upon work supported by Robert Bosch LLC. We gratefully acknowledge the helpful comments received from anonymous reviewers.

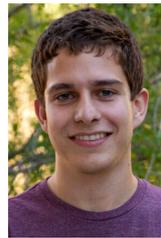

**Jeremy Morton** is a Ph.D. candidate in the Department of Aeronautics and Astronautics at Stanford University, and a member of the Stanford Intelligent Systems Laboratory. He received a B.S. in Aerospace Engineering from the University of Illinois at Urbana-Champaign and an M.S. in Aeronautics and Astronautics from Stanford University. His research interests include deep learning, reinforcement learning, and validation of autonomous driving systems.

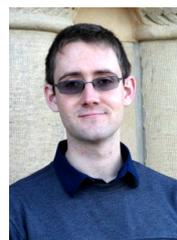

**Tim A. Wheeler** is a Ph.D. candidate in the Department of Aeronautics and Astronautics at Stanford University, and a member of the Stanford Intelligent Systems Laboratory. He received an M.S. in Aeronautics and Astronautics from Stanford. His research focuses on the validation and optimization of active automotive safety systems.

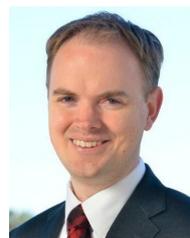

**Dr. Mykel Kochenderfer** is Assistant Professor of Aeronautics and Astronautics at Stanford University. He received his Ph.D. from the University of Edinburgh and B.S. and M.S. degrees in computer science from Stanford University. Prof. Kochenderfer is the director of the Stanford Intelligent Systems Laboratory (SISL), conducting research on advanced algorithms and analytical methods for the design of robust decision making systems.